# The Nipple-Areola Complex for Criminal Identification

[a] Wojciech Michal Matkowski*, [b] Krzysztof Matkowski, [a] Adams Wai-Kin Kong*, [c] Cory Lloyd Hall
[a] School of Computer Science and Engineering, Nanyang Technological University, Singapore
[b] Faculty of Medicine, Medical University, Wroclaw
[c] The MITRE Corporation, Washington, D.C.

## Abstract

*In digital and multimedia forensics, identification of child sexual offenders based on digital evidence images is highly challenging due to the fact that the offender's face or other obvious characteristics such as tattoos are occluded, covered, or not visible at all. Nevertheless, other naked body parts, e.g., chest are still visible. Some researchers proposed skin marks, skin texture, vein or androgenic hair patterns for criminal and victim identification. There are no available studies of nipple-areola complex (NAC) for offender identification. In this paper, we present a study of offender identification based on the NAC, and we present NTU-Nipple-v1 dataset, which contains 2732 images of 428 different male nipple-areolae. Popular deep learning and hand-crafted recognition methods are evaluated on the provided dataset. The results indicate that the NAC can be a useful characteristic for offender identification.*

## 1. Introduction

Traditionally, in the investigation of child sexual assault, the investigators use biometric characteristics such as latent prints, or DNA from, e.g., body fluids collected at a crime scene [1]. Nonetheless, sometimes the crime scene location may remain unknown or the crime is unreported making the physical examination and collection of such evidence impossible. Nowadays, more evidence is also in the form of digital images, because the offenders take pictures for their personal record or to share with other pedophiles, e.g., on Dark Web. The offenders cover or hide their faces, to avoid personal identification and prosecution when forensic investigators or law enforcement agencies take over the child sexual assault materials. Some researchers propose to use skin marks [2], skin texture [3], visualized veins [4] or androgenic hair [5] to identify criminals and victims in images where no obvious traits such as face or tattoos are visible. Each of them has its own limitations, e.g., not enough skin marks or hair, no clear veins, etc. make the identification challenging. Regardless of the availability of these characteristics, other ones can be used to support the evidence or provide some useful clues for the investigator. In this paper, the male nipple-areola complex (NAC) is proposed for criminal identification. The offender's chest and thus NAC can be visible in some of the sexual assault images.

In the computer vision related literature, studies on nipples mostly aim nipple detection in color images for pornographic/nude image detection [6], [7] or nipple detection in medical images [8], [9]. According to our best knowledge, no one studies the NAC identification for forensic investigation.

The rest of this paper is organized as follows. In Section 2, basic information about the NAC morphology and related medical studies on the NAC properties are given. In Section 3, the new NTU-Nipple-v1 dataset is described. In Section 4, the experiments are performed. In Section 5, the conclusions are given.

## 2. Nipple-areola complex

NAC contains nipple (papilla), areola and tubercles. Fig. 1 shows male NAC. The nipple is located approximately in the center of the NAC and surrounded by the areola. The areola's color is darker due to melanin accumulation and its structure is uneven due to the presence of tubercles [10], [11]. The NAC, vary in shape, dimension, texture and color among different people. In addition, two NAC of the same person also vary in appearance [12]. In some cases of, e.g., breast surgery, massive weight loss, trauma, gender reassignment and gynecomastia affect the NAC's morphology. The most information about the NAC visual appearance is found in the reconstructive and aesthetic surgery literature. The measurement studies focus on a position and configuration of the male NAC in the chest and its shape and dimensions [13], [14], [15], [16]. Researchers also study the correlation between male NAC parameters and body weight, height, body mass index (BMI) [13], [16] and the size of pectoral muscle [15]. However, the medical literature mostly focuses on female rather than male NAC [13].

*Corresponding authors: W.M. Matkowski, email: matk0001@e.ntu.edu.sg
A.W.K. Kong, email: adamskong@ntu.edu.sg

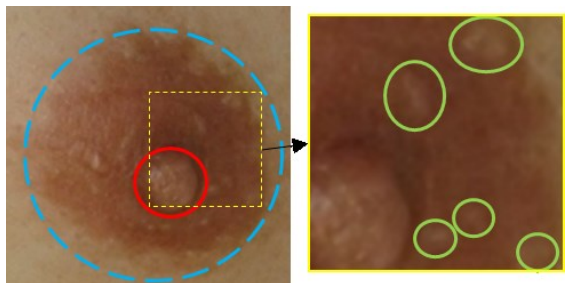

Figure 1: Left: Nipple-areola complex with approximately marked nipple (red solid), areola (blue dashed). Right: close-up image with marked tubercles (green solid).

## 3. NTU-Nipple-v1 dataset

NTU-Nipple-v1 dataset was collected during two different sessions with at least a one-week time interval between sessions. The images were taken by Canon EOS 500D or NIKON D70s cameras. There were no strict posing requirements for subjects. The subjects are male and mainly Chinese, Indian, Malay and some Caucasian. The subjects' NAC are cropped using a square bounding box such that the nipple is approximately in the center of the image (see Fig. 3). The dataset consists of 2732 images from 428 different NAC. Fig. 2 shows examples of the NAC images from the dataset. Note, that the right and left NAC from the same subject are considered as a different NAC. The median NAC image size is 179 by 179 pixels and the median number of images of a single NAC is 6. Fig. 4 shows some quantitative statistics of the dataset. NTU-Nipple-v1 dataset will be available online [17] in three months after this paper is published.

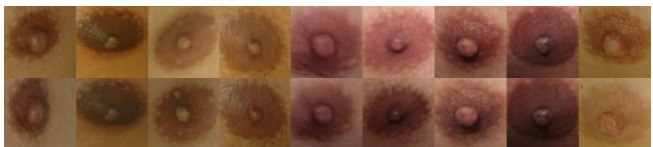

Figure 2: Examples of different NAC from NTU-Nipple-v1 dataset from session 1 (top) and session 2 (bottom). Images in the same column are from the same subject's NAC.

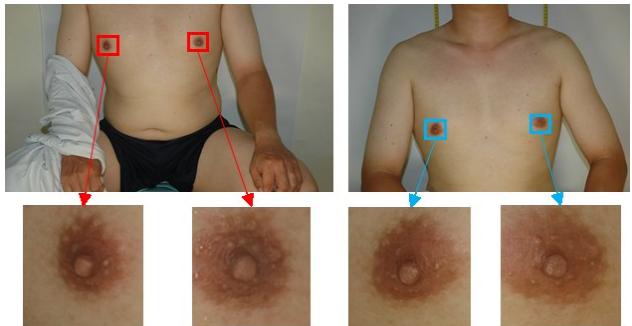

Figure 3: Examples of cropped NAC images. NAC are highlighted in the original chest images. The NAC are from the same subject from two different sessions.

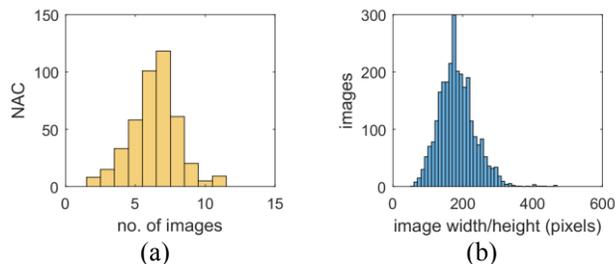

(a)            (b)

Figure 4: Histograms showing (a) the distribution of images of the same single NAC, (b) the width/height of NAC images.

## 4. Experiments

NTU-Nipple-v1 dataset is split into training/gallery and testing/probe sets. The images in the gallery and probe are from different sessions. Each probe NAC has its corresponding one in the gallery and vice versa. The gallery and the probe sets consist of 1577 and 1155 NAC images, respectively. The experiments in the NAC identification (1-to-N comparisons) scenario are performed. The results are presented in cumulative match characteristic (CMC) curves. CMC is a common evaluation metric for forensic applications. In the presented results, the CMC's ranks range from 1 to 30. The human expert may visually examine, e.g., top 30 ranks, which is a common practice in forensic investigation, e.g., in latent prints identification [18]. Therefore, not only rank-1 but also higher ranks such as rank-30 are important.

### 4.1. Recognition algorithms

Five popular deep learning architectures AlexNet [19], VGG-16 [20], VGG-Face [21], GoogLeNet [22] and ResNet-50 [23] are selected for evaluation. All these architectures except VGG-Face were pretrained on the ImageNet dataset. The VGG-Face was pretrained on the face image dataset [21]. The pretrained models and fine-tuning approach are considered because NTU-Nipple-v1 dataset is relatively small and training from scratch is likely to overfit on such dataset. In addition, two popular hand-crafted features local binary pattern (LBP) with the matching method from [24] and scale invariant feature transform (SIFT) with L2 feature matching are also used in the evaluation.

### 4.2. Implementation details

The deep learning algorithms are implemented using MatConvNet library and the pretrained models [25]. These pretrained models require fixed sized input images. Thus NAC images are resized to 224 by 224 pixels (227 by 227 for AlexNet). The last softmax layer is fine-tuned (Section 4.5) using ADAM [26] optimizer with 0.001 learning rate, 0.0005 weight decay, 0.9 momentum and 128 batch size. When fine-tuning lower layers (Section 4.6) their learning rate is set to 10 times smaller, which is 0.0001. For hand-crafted features (Section 4.7), the color images are first transformed into the grayscale. Uniform LBP image is used with radius 2, 8 neighbors and the LBP histograms are extracted in a 7 by 7 uniform grid and matched using Chi-square distance. The SIFT descriptors are detected using VL_FEAT library [27] with the threshold set to 100 and matched with the threshold set to 1.6.

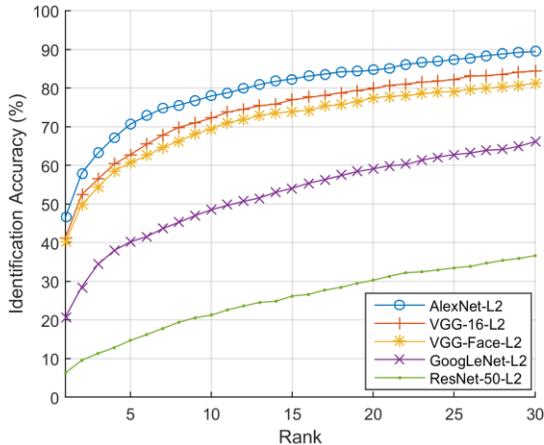

Figure 5: CMC curve of different deep learning architectures where L2 distance is used for identification.

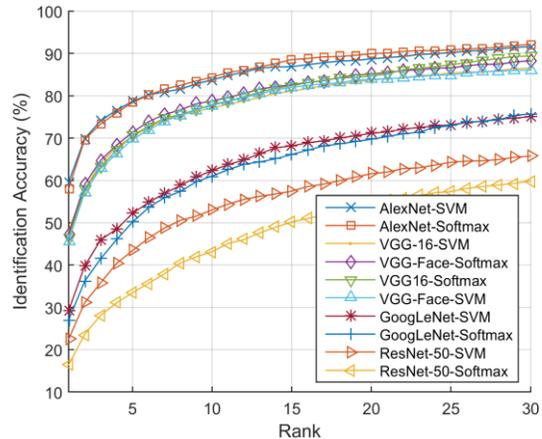

Figure 6: CMC curve of the deep learning algorithms where softmax and SVM are used for identification.

### 4.3. Deep features

In this experiment, the deep features extracted from the one before the last layer are evaluated. The feature vectors are L2-normalized and the L2 distance is used to obtain comparison scores. The feature vector sizes for AlexNet, VGG-16, VGG-Face, GoogLeNet and ResNet-50 are 4096, 4096, 4096, 2048, 1024, respectively. The results are given in Fig. 5. AlexNet, which is the "shallowest" among the selected architectures, achieves the highest performance. Its rank-1 and rank-30 accuracies are 46.58% and 89.44%, respectively.

### 4.4. Fine-tuning the last layer

In the previous Section 4.3, the 1-nearest neighbor in the Euclidean space (L2) is used for identification. In this experiment, the softmax and linear support vector machine (SVM) are employed to investigate the parametric classifiers. SVM is used in the one-against-all setting, the feature vectors are standardized (z-score) before training, and the positive labels are set to 1 and the negative ones to -1. The results are presented in Fig. 6. For all five deep learning architectures, the performance increases when using softmax or SVM comparing to the performance in Section 4.3. The order of the five deep architectures with respect to accuracy remains the same as in Section 4.3. In addition, SVM generally achieves higher results. The rank-1, rank-15 and rank-30 accuracy are given in Table 1.

Table 1: The rank accuracies (%) of the deep learning architectures using softmax and SVM for identification.

| Architecture | Classifier | Rank-1 | Rank-15 | Rank-30 |
|---|---|---|---|---|
| AlexNet | SVM | **59.57** | 86.92 | 91.60 |
| AlexNet | Softmax | 58.00 | **88.57** | **92.12** |
| VGG-16 | SVM | 48.31 | 81.12 | 86.49 |
| VGG-16 | Softmax | 46.58 | 82.33 | 89.43 |
| VGG-Face | SVM | 45.71 | 81.88 | 86.06 |
| VGG-Face | Softmax | 47.27 | 82.68 | 88.31 |
| GoogLeNet | SVM | 29.26 | 68.13 | 75.06 |
| GoogLeNet | Softmax | 27.01 | 66.14 | 75.75 |
| ResNet-50 | SVM | 22.51 | 57.49 | 65.80 |
| ResNet-50 | Softmax | 16.36 | 50.21 | 59.82 |

### 4.5. Fine-tuning fully connected layers of AlexNet

In this experiment, the aim is to fine-tune the fully connected layers. AlexNet architecture is selected because it achieves the highest performance in previous experiments in Section 4.3 and 4.4. To avoid, fast overfitting, the dropout layer with the dropout rate set to 0.5 is put before the last fully connected layer and the data augmentation on the fly is used. For augmentation, the contrast and the saturation random shifts are applied with the shift factor set to 0.5. Additionally, aggressive spatial random vertical and horizontal translations and scaling are applied. To avoid bordering effects, the 1.3 bigger NAC image is translated and scaled and then cropped as shown in Fig. 7. The crop is always within the image boundaries. The results are presented in Fig. 8. In the figure legend, the "FTFC" indicates the results obtained in this experiment. The highest rank-1 accuracy of 62.16% is achieved by FTFC-AlexNet-L2. This result indicates that the employed data augmentation can boost performance. Especially, when 1-nearest neighbor L2 is used for identification, the rank-1 accuracy increases by 15.58% (AlexNet-L2 vs FCFT-AlexNet-L2).

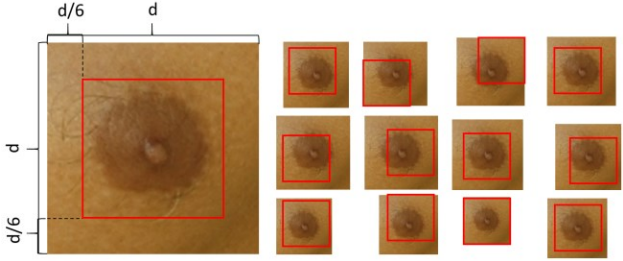

Figure 7: Schematic illustration of the spatial data augmentation. During the training, the bigger NAC image is randomly cropped within the red box and input to the network.

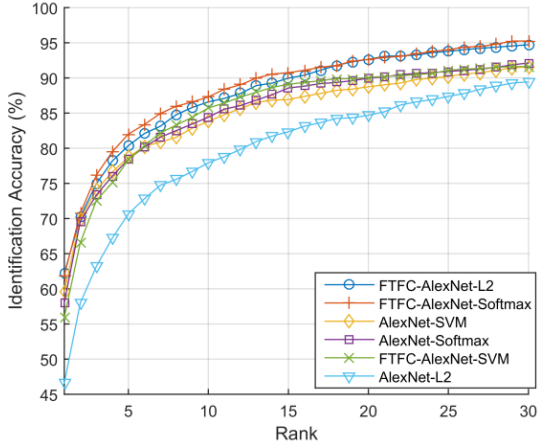

Figure 8: CMC curve comparing AlexNet (Section 4.3 and 4.4) and the AlexNet with fully connected layers fine-tuned and the augmentation applied (Section 4.5).

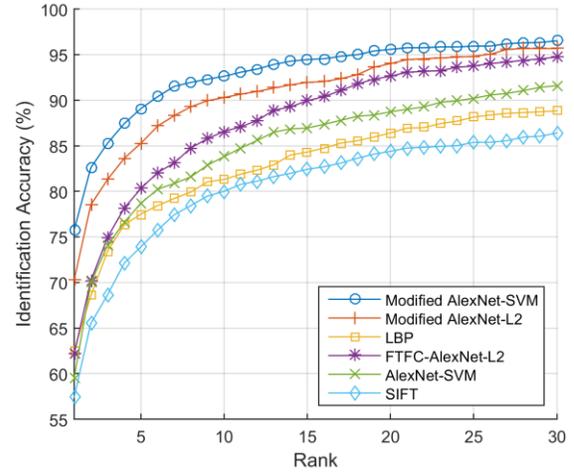

Figure 9: CMC curve comparing the best AlexNet from Sections 4.4 and 4.5 and the Modified AlexNet (Section 4.6), and hand-crafted LBP and SIFT methods.

Table 2: The rank accuracies (%) of the AlexNet architectures and the hand-crafted methods.

| Method | Rank-1 | Rank-15 | Rank-30 |
|---:|---:|---:|---:|
| Modified AlexNet-SVM | **75.76** | **94.46** | **96.54** |
| Modified AlexNet-L2 | 70.30 | 91.95 | 95.67 |
| LBP | 62.51 | 84.33 | 94.72 |
| FTFC-AlexNet-L2 | 62.16 | 89.96 | 86.49 |
| AlexNet-SVM | 59.57 | 86.93 | 91.60 |
| SIFT | 57.40 | 82.42 | 86.41 |

### 4.6. Modified fully connected layers of AlexNet

In this experiment, the comparison of deep learning and hand-crafted based methods is given. The hand-crafted LBP and SIFT are used as described in Section 4.2. LBP and SIFT achieve 62.51% and 57.5% rank-1 accuracy, respectively. These results are higher than all deep learning results from Section 4.3. Note that, LBP and SIFT also use the same type of classifier (L2 distance, 1-nearest neighbor) as is used in Section 4.3. In addition, LBP outperforms at rank-1 all the results from Section 4.4, where the SVM and softmax classifier are used for identification. SIFT performance at rank-1 is only worse than AlexNet. These results indicate that more specialized deep learning architectures and training techniques are needed to increase the identification accuracy.

The additional experiment, using AlexNet is performed to give some insights and further investigate the possibility of boosting deep learning performance. The AlexNet's fully connected layers (which contain over 54 million parameters) are replaced with one fully connected (4.7 million parameters) layer with 512 neurons and softmax layer, with two dropouts put in-between with 0.5 dropout rate. The same augmentation as in Section 4.5 is used. The results are presented in Fig. 9 and Table 2. Modified AlexNet and L2 distance are able to achieve the highest performance among the selected methods. It achieves 70.2% and 95.67% rank-1 and rank-30 accuracy, respectively. Employing SVM further increases the identification performance to 75.76% and 96.54%, rank-1 and rank-30 accuracy, respectively. Fig. 10 shows the examples of top-10 matches returned by the Modified AlexNet-SVM.

### 5. Conclusions

Criminal identification in the evidence images with no obvious characteristic such as face or tattoos is challenging in forensic investigation. In this paper, a single NAC is proposed to identify child sexual offenders in the images where chest and NAC are visible. NTU-Nipple-v1 dataset which contains 428 different male NAC is provided. Five popular deep learning architectures and two well-known hand-crafted methods are employed in the evaluation. The experimental results indicate that the NAC can be a useful characteristic for forensic applications. In the experiments, the highest reported rank-1, rank-15 and rank-30 identification accuracies are 75.76%, 94.46% and 96.54%, respectively. To further increase the accuracy, more research effort and specialized algorithm development are needed. Additionally, in the future, a score fusion of two NAC from the same subject, NAC detection and alignment could also be investigated. More NAC images from an uncontrolled environment, with low-resolution, large pose and illumination variations should be collected to evaluate algorithms in more challenging scenarios.

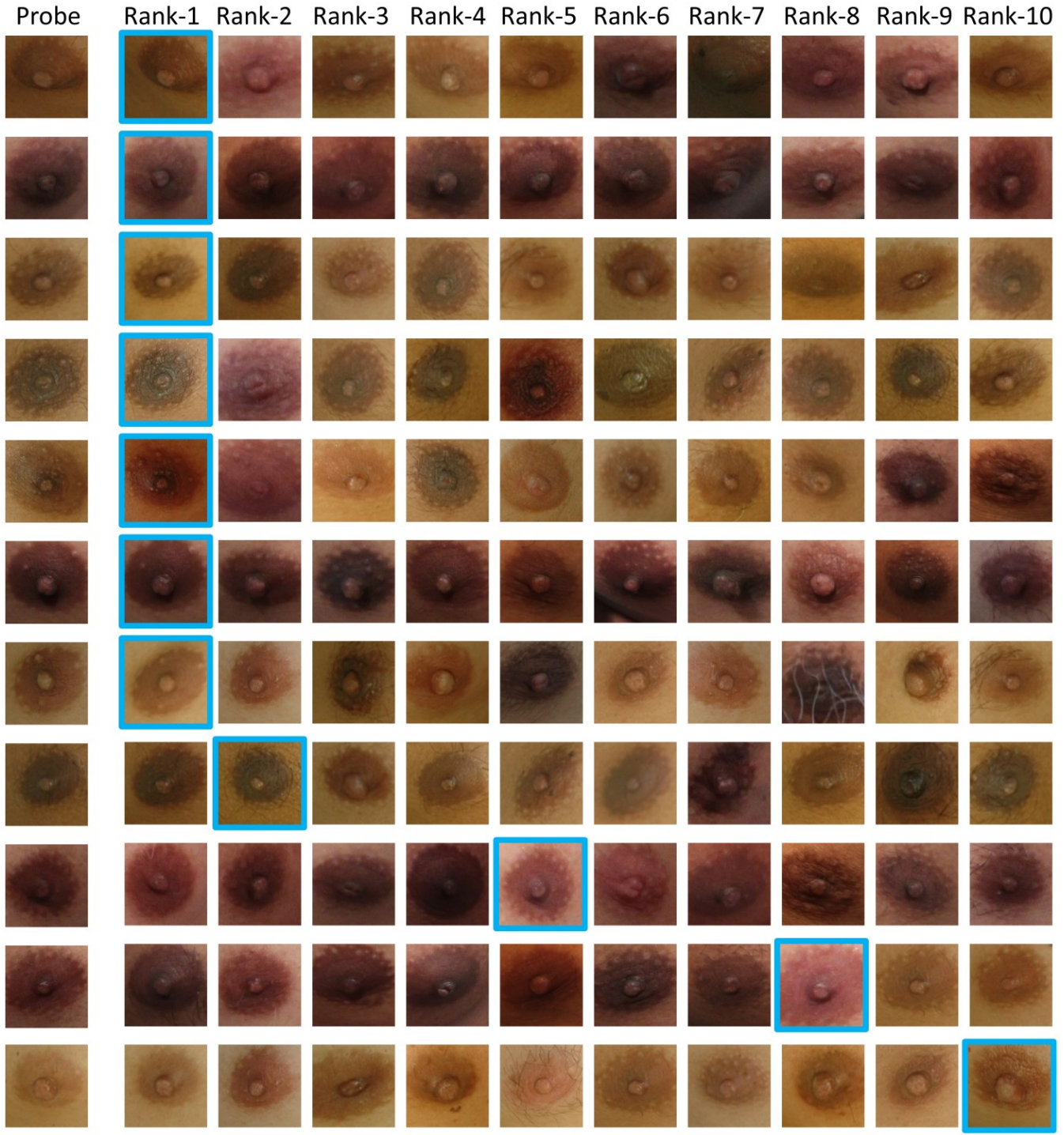

Figure 10: Rows: examples of top-10 identification results using Modified AlexNet-SVM. The true matches between each probe and the rank-ordered gallery images are in the blue boxes.

## Acknowledgment

This work is partially supported by the Ministry of Education, Singapore through Academic Research Fund Tier 2, MOE2016-T2-1-042(S).